\def\year{2019}\relax
\documentclass[letterpaper]{article} 
\usepackage{aaai19}  
\usepackage{times}  
\usepackage{helvet}  
\usepackage{courier}  
\usepackage{url}  
\usepackage{graphicx}  
\usepackage{enumitem}
\usepackage{subfigure}
\usepackage{url}
\usepackage{color}
\usepackage{multirow}
\usepackage{array}
\usepackage[linesnumbered,ruled,vlined]{algorithm2e}
\usepackage{algorithmic}
\usepackage{amsmath,amsfonts,amssymb,bbm,epsfig,bm}
\def\etal{{\em et al.\/}\,}
\newcommand{\seq}[1]{\mathbf{#1}}

\begin{document}
%
\title{Generating Multiple Diverse Responses for Short-Text Conversation}

 \author{Jun Gao\textsuperscript{1}\thanks{Work done when Jun Gao was interning at Tencent AI Lab.},
Wei Bi\textsuperscript{2}\thanks{Corresponding author},
Xiaojiang Liu\textsuperscript{2},
Junhui Li\textsuperscript{1},
Shuming Shi\textsuperscript{2}\\
\textsuperscript{1}{School of Computer Science and Technology, Soochow University, Suzhou, China}\\
imgaojun@gmail.com, lijunhui@suda.edu.cn\\
\textsuperscript{2}{Tencent AI Lab, Shenzhen, China}\\
{\{victoriabi, kieranliu, shumingshi\}@tencent.com}
}
\maketitle
\begin{abstract}
Neural generative models have become popular and achieved promising performance on short-text conversation tasks. They are generally trained to build a 1-to-1 mapping from the input post to its output response.
However, a given post is often associated with multiple replies simultaneously in real applications.
Previous research on this task mainly focuses on improving the relevance and informativeness of the top one generated response for each post. Very few works study generating multiple accurate and diverse responses for the same post.
In this paper, we propose a novel response generation model, which considers a set of responses jointly and generates multiple diverse responses simultaneously.
A reinforcement learning algorithm is designed to solve our model.
Experiments on two short-text conversation tasks validate that the multiple responses generated by our model obtain higher quality and larger diversity compared with various state-of-the-art generative models.
\end{abstract}

\section{Introduction}
\noindent
Endowing the machine with the ability to converse with humans using natural language is one of the fundamental challenges in artificial intelligence~\cite{turing1950}. 
In particular, conversation models in open domains have received increasing attention due to its wide applications including chatbots, virtual personal assistants and etc. 
With the vast amount of conversation data available on web services,
generative models have shown their great potential.
Especially, the Sequence-to-sequence~(Seq2seq) framework ~\cite{vinyals2015neural,sordoni2015neural}, which learns the mapping from the input to its corresponding output directly,  has achieved promising performance.
%
%
%
Following the most conventional setting for generative short-text conversation models~\cite{shang2015neural,li2016simple,mo2016personalizing,shen2017conditional}, we consider the single round chi-chat conversation with no context information, i.e. an input
post from a user and the output response given by
the machine.



Despite the popularity of the Seq2seq models, various problems occur when they are applied for short-text conversation.
One important issue is that they try to model the input post and the output response as a 1-to-1 mapping.
Whereas, a post is often associated with multiple responses simultaneously in real applications, which is a 1-to-n mapping.
For example, for a user's tweet on Twitter, there are often multiple replies obtained from the other users.

In order to handle a set of replies, a simple method is to pair each of them with the input as multiple training instances and then train the Seq2seq models. 
Previous results show that this method tends to generate generic responses~\cite{li2016diversity,li2016deep}.
Moreover, only one target response can be generated in this method.
To obtain multiple responses as in the real applications,  many approaches 
resort to beam search or enhanced beam search for diverse answers~\cite{li2016simple,shao2017generating}. 
%
Recently, Zhou \etal~\shortcite{zhou2017mechanism,zhou2018elastic} introduced multiple latent embeddings as language responding mechanisms into the Seq2seq framework. By choosing different responding mechanisms, multiple output responses can be generated. 
Zhao \etal~\shortcite{zhao2017learning} adopted the conditional variational auto-encoder (CVAE) and generated multiple responses by drawing samples from the prior distribution.
Yet, these methods still decomposed multiple responses into independent ones, and did not model the fact of multiple responses.


In this paper, we propose a novel response generation model for short-text conversation, which considers multiple responses jointly.
We introduce a latent variable distribution for each post and sample different values from the learned latent distribution to generate multiple diverse responses.
Our novelty lies in two aspects. 
First, we model the post with its various responses as a bag of instances and infer the distribution of the latent variable by considering multiple responses jointly. 
Second, unlike in CVAE, we adopt interpretable latent variables, of which different samplings should be capable to capture the divergence between different responses.
Here, we set the latent variable as a word in the vocabulary. Each sampling of the latent variable is then to select a specific word, which drives the generated response to be more meaningful around this word.

Since open-domain conversations cover a wide range of topics and areas, we require a large vocabulary 
to adequately include the possible latent words in all responses. 
Thus, it is impractical to compute the exact gradient from the fully factorized probabilities over all latent variables and apply the standard back-propagation training procedure.
We propose to treat each latent word as an action, and use a reinforcement learning(RL) algorithm to solve the large latent space issue. We first use the latent word inference network to estimate the distribution over all words for an input post.
Based on this probability distribution, we sample multiple latent words, which are then combined as parts of inputs in the generation network.
The generation network will return the reward of each sampled latent words by considering all generated responses jointly.
Specifically, we design a sampling scheme to select multiple diverse latent words for our RL algorithm to converge effectively.


To summarize, our contributions are threefold:
\begin{enumerate}[wide=0.5\parindent]
\item We propose a response generation model which directly considers an input post with its multiple responses jointly. Our model could automatically generate multiple diverse responses for a single post.
\item We consider latent words as actions and propose to use a reinforcement learning algorithm to solve our model.
\item Experimental results show that our model outperforms existing state-of-the art generative methods in generating multiple high-quality and diverse responses. All our code and datasets are available are \url{https://ai.tencent.com/ailab/nlp/dialogue.html}.
\end{enumerate}



\section{Related Work}
\label{sec:related}
The Seq2seq framework has been widely used for conversational response generation~\cite{vinyals2015neural,sordoni2015neural,shang2015neural}.
Such models learn the mapping from an input $\seq{x}$ to one output $\seq{y}$ by maximizing the pairwise probability of $p(\seq{y}|\seq{x})$.
During testing, these models only target for one response.
In order to obtain multiple responses, beam search can be used.
However, the resulting multiple sequences are often very similar.
Many approaches have been proposed to re-rank diverse meaningful answers into higher positions.
For example,
 Li \etal~(\citeyear{li2016simple}) proposed a simple fast decoding algorithm to directly encourage response diversity in the scoring function used in beam search.
Shao \etal (\citeyear{shao2017generating}) heuristically
re-ranked the responses segment by segment to inject diversity earlier in the decoding process. 
These methods only modified the decoding steps and still often generated responses using different words but with similar semantics.

A few works have explored different factors that decide the generation of diverse responses.
For task-oriented dialogue systems,  Wen \etal ~\shortcite{wen2017latent} proposed to model a latent space to represent intentions.
Zhao \etal~\shortcite{zhao2017learning}
adopted the CVAE for learning discourse-level diversity for dialog models. 
They further incorporated a distribution over potential dialogue acts for better discourse-level diversity.
Unlike them, our work targets for the single round open-domain conversation task, which assumes no discourse level information.
Also, it is observed that in CVAE with a fixed Gaussian prior, the learned conditional posteriors tend to collapse to a single mode, yielding little diversity in the generated results~\cite{wang2017diverse}. In our model, if we choose two latent words with far different meanings, the generated responses should be  different.

The most relevant work to ours is from Zhou \etal ~\shortcite{zhou2017mechanism,zhou2018elastic}.
They 
introduced latent embeddings as the language responding mechanisms into the Seq2seq model.
By choosing different responding mechanisms, multiple responses can be generated. 
However,  
their model decomposed the joint probability of multiple responses into a product of individual probabilities:
$p(\{\seq{y}\}|\seq{x}) = \prod_{\seq{y}} p(\seq{y}|\seq{x})$.
Despite that the latent embedding has no interpretable meaning, they have to set a very small number of mechanisms
in order to optimize the fully factorized probabilities over the latent embeddings in a standard back-propagation training procedure. 
Our model assumes a much larger latent space with interpret ability.
Thus, it is impractical to compute the exact gradient, which we will show in later section. We then propose a reinforcement learning algorithm to train our model.

We notice that Yao \etal~\shortcite{yao2017towards} proposed
to generate an informative reply based on a cue word.
In our model, we utilize a latent word to guide the response generation, which shares certain similarities. 
Their results also validates our model assumption that it should be effective to integrate auxiliary word-level information for generating high-quality responses. 
However, they simply computed the cue word for each post using the point-wise mutual information (PMI) from the word concurrence statistics, and focused on designing a better decoding cell to deal with the cue word.
Our work considers another problem, which is to infer multiple diverse latent words from a parametric model to generate multiple responses with satisfying quality and diversity. 

Most of the related methods mentioned above will be compared in our experiments and our results show that our proposed method outperforms them in terms of generating multiple relevant and diverse responses.

\section{Models}
\subsection{Problem Formulation and Model Overview}
\label{sec:overview}
We are given training samples $\{(\seq{x},\{\seq{y}\})\}$, with $\seq{x}$ denoting the input post
and $\{\seq{y}\}$ its set of output responses.
We assume an unobserved latent variable $z$ for each input $\seq{x}$. 
Before introducing our model, we first discuss how to construct a proper latent space $\seq{Z}$ in short-text conversation tasks:

\begin{itemize}[wide=0.5\parindent,noitemsep]
\item For any $z \in \seq{Z}$, it should be interpretable, which can show its relevance to the input $\seq{x}$ as well as the responses $\{\seq{y}\}$. Thus, we are easy to identify whether a good unobservable $z$ is assigned for a sample $(\seq{x}, \{\seq{y}\})$.
\item Each $z \in \seq{Z}$ should be capable of capturing the sentence-level discrepancy such that given distinct $\seq{x}$'s, the inferred $z$'s can be different and given distinct $z$'s, the generated responses can be different.
\end{itemize}
To meet these two criteria, we set the latent space $\seq{Z}$ as the vocabulary used in each task and each $z$ corresponds to one of its words. In terms of the first criterion,  we can directly check the relevance from the plain text of $z$, $\seq{x}$ and $\{\seq{y}\}$. For the second one, the vocabulary defines a large enough latent space for a given $\seq{x}$ to select out multiple distinct and diverse $z$'s to generate multiple responses.

We observe that it is often the case that not all responses relate to one single latent word $z$. 
For example, with a query ``I feel boring in the weekends. How do you spend your time?'', it has two responses ``I play guitar sometimes'' and ``My favorite is camping and you can join me next time.''.
Both ``guitar'' and ``camping'' are valid latent words of the query, but each of them is relevant to only one of the responses.
Thus, we get inspired from multi-instance learning~\cite{zhou2004multi}, in which a positive bag of instances contains at least one positive instance. 
Here, a query with its multiple responses is considered as a bag of instances. That is, if a given $\seq{x}$ is assigned with a latent word $z$, its bag of instances contains at least one response $\seq{y}$ relevant to this $z$. In turn, if multiple responses are relevant to different $z$'s, multiple $z$'s can be inferred with relatively large probabilities simultaneously for a single $\seq{x}$.
Our training objective is to minimize the average risk of generating the bag of responses $\{\seq{y}\}$ over the probability distribution of $z$ given $\seq{x}$, which is defined as:
\begin{equation}
 J(\seq{\Theta}) = \mathcal{L}(\{\seq{y}\} | \seq{x}) \
= \mathbb{E}_{p(z|\seq{x})} [\mathcal{L}(\{\seq{y}\} | \seq{x}, z)], \label{eq:factorize}
\end{equation}
where $p(z|\seq{x})$ is the probability of a latent $z$ given $\seq{x}$,
and $\mathcal{L}(\{\seq{y}\} | \seq{x}, z)$ is the loss of generating the bag of responses $\{\seq{y}\}$ given both $\seq{x}$ and $z$.
In our model, we use deep neural networks to parameterize $p(z|\seq{x})$
and $\mathcal{L}(\{\seq{y}\} | \seq{x}, z)$, denoted as $\seq{W}$ and $\seq{G}$ respectively. $\seq{\Theta}=\{\seq{W}, \seq{G}\}$ contains all model parameters. Note that  
unlike previous works~\cite{zhou2017mechanism,zhou2018elastic}, we do not decompose $\{\seq{y}\}$ into individual $\seq{y}$'s and infer the latent $z$ for each of them separately.

\begin{figure}[tb]
\begin{center}
 \includegraphics[width=\linewidth]{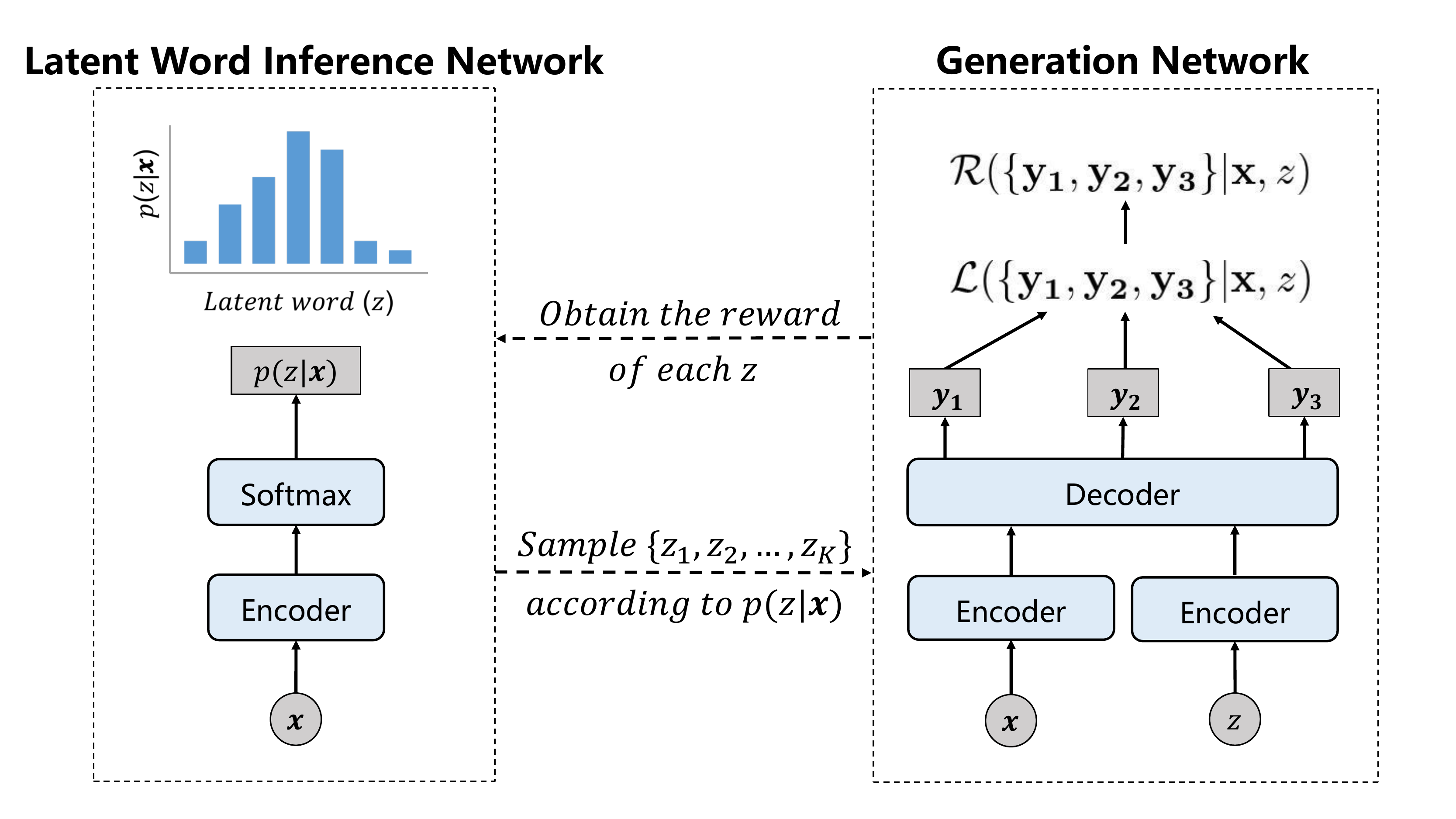}
\end{center}
\caption{An overview of the proposed model. The given $\seq{x}$ is associated with three output responses $\{\seq{y}_1,\seq{y}_2,\seq{y}_3\}$. Each latent $z$ is a word in the vocabulary.}
\label{fig:rl}
\end{figure}

The overall framework of our model is shown in Fig.~\ref{fig:rl}.
It consists of two key components:
a latent word inference network for $p(z|\seq{x})$, from which multiple words can be sampled out to drive the generation of diverse responses; 
a response
generation network for $\mathcal{L}(\{\seq{y}\} | \seq{x}, z)$, 
which generates the given multiple responses according to the input $\seq{x}$ and sampled words.
We iteratively update the two networks in that:
  the latent word inference network starts with an initial policy, i.e. $p(z|\seq{x})$,  revise its strategy based on the reward and improve the accuracy of the inferred latent words; 
the generation network receives more accurate latent words as inputs to facilitate generating the given responses, and output the reward for each sampled $z$ for updating the latent word inference network.  
In the following  subsections, we describe these two networks,  the RL algorithm for their joint training and prediction in detail.



\subsection{Latent Word Inference Network}
This network estimates the probability distribution $p(z|\seq{x})$.
%
%
%
%
  We first encode the input $\seq{x}$ with a bidirectional GRU 
  to obtain an input representation $\seq{h}_{\seq{x}}$
  and then compute the probability of each latent word as follows:
   \begin{eqnarray}
  p(z| \seq{x})  =  \mbox{softmax}({\seq{W}_2}
  \cdot \mbox{tanh}( \seq{W}_1 \seq{h}_{\seq{x}} + \seq{b}_1 ) +  \seq{b}_2), \label{eq:softmax}
 \end{eqnarray}
 where parameters in the bidirectional GRU, $\seq{W}_1$, $\seq{b}_1$, $\seq{W}_2$ and $\seq{b}_2$ are jointly trained in our model. 

 \subsection{Generation Network}
We assume that $\mathcal{L}(\{\seq{y}\}|\seq{x}, z)$ can be estimated from individual loss $\ell(\seq{y}|\seq{x}, z)$'s for $\seq{y} \in \{\seq{y}\}$ through a simple parameter-free and differentiable function $f(\cdot)$:
\begin{eqnarray}
\mathcal{L}(\{\seq{y}\}|\seq{x}, z) = f_{\seq{y} \in \{y\}}(\ell(\seq{y}|\seq{x}, z)).
\end{eqnarray}
We first introduce the detailed network structure to estimate $\ell(\seq{y}|\seq{x}, z)$, and then discuss proper choices of  $f(\cdot)$.

With a sampled $z$, we first encode $\seq{x}$ through another bidirectional GRU for an input representation $\seq{h}_{\seq{x}}^g$.
At the same time, we obtain the representation $\seq{h}_{z}^g$ of $z$ 
%
using an embedding layer followed by a fully connected layer.
Both representations are then leveraged to decode the output sequence $\seq{y}$. 
Here, our decoding function is constructed as follows. We compute the attention between $\seq{h}_{z}^g$ and $\seq{x}$ and construct the context vector $\seq{c}_z$: 
\begin{eqnarray}
a_{j} = \frac{\exp({\seq{h}_{z}^g}^T \seq{h}_{\seq{x}}^g(j))}{\sum_{t}\exp({\seq{h}_{z}^g}^T \seq{h}_{\seq{x}}^g(t))},\;\;
	\seq{c}_{z} = \sum_{t} a_{t}\seq{h}_{\seq{x}}^g(t), \label{eq:attn_2}
\end{eqnarray}
where $\{{\seq{h}_{\seq{x}}^g(t)}\}$ is the series of hidden vectors obtained in the encoder for each word in $\seq{x}$.
Intuitively, $\seq{c}_{z}$ is a weighted sum of the input representations, which captures the information of $\seq{x}$ relevant to the selected $z$.

%
We use GRU as the decoding cell with $\{\seq{h}_{\seq{y}}(t)\}$ denoting its series of hidden vectors. In each decoding step, we calculate the attention and the context vector $\seq{c}_{\seq{y}}(t)$ for $\seq{h}_{\seq{y}}(t)$, which can be obtained by replacing $\seq{h}_z^g$ with $\seq{h}_{\seq{y}}(t)$ in Eq.~\ref{eq:attn_2}. 
Then, we concatenate $\seq{h}_{\seq{y}}(t)$, $\seq{c}_{z}$ and $\seq{c}_{\seq{y}}(t)$  together and forward this vector into a fully connected layer followed by a softmax output layer. $\ell(\seq{y}|\seq{x}, z)$ is set as the negative log likelihood of $p(\seq{y}|\seq{x}, z)$.
Comparing with the decoding step in traditional Seq2seq models~\cite{bahdanau2015neural}, we integrate more information from the input representations relevant to the selected word $z$ to help predict the current output word $\seq{y}(t)$.





Next, we discuss how to design a proper function $f(\cdot)$ with $\ell(\seq{y}|\seq{x},\seq{z})$'s estimated from the above network structure. Typically, 
$f(\cdot)$ is set to be the average function:
\begin{eqnarray}
\mathcal{L}_{avg}(\{\seq{y}\}| \seq{x}, z) = \frac{1}{|\{\seq{y}\}|}\sum_{\seq{y} \in \{\seq{y}\}} \ell(\seq{y}|\seq{x}, z), \label{eq:sum}
\end{eqnarray}
where $|\{\seq{y}\}|$ is the cardinality of $\{\seq{y}\}$.
However  $\mathcal{L}_{avg}$ may not be proper in our setting.
As discussed in Section~\ref{sec:overview}, not all responses may relate to a single $z$.
Thus,  the loss terms $ \ell(\seq{y}|\seq{x}, z)$'s for those $\seq{y}$'s relevant to $z$ will be small, whereas the remaining will be large.
The overall loss
$\mathcal{L}_{avg}(\{\seq{y}\}| \seq{x}, z)$ could still be large. 
Hence, we propose to use the minimum function as the $f(\cdot)$ function:
\begin{eqnarray}
\mathcal{L}_{min}(\{\seq{y}\}| \seq{x}, z) = \min_{\seq{y} \in \{\seq{y}\}} \ell(\seq{y}|\seq{x}, z). \label{eq:min}
\end{eqnarray}
If the sampled $z$ paired with $\seq{x}$ can generate one of its responses very well, its overall loss value now will be small.

\subsection{Joint Training with Reinforcement Learning}
To update the networks, we generally compute the
gradient of $ J(\seq{\Theta})$ in Eq.~\ref{eq:factorize} with respect to $\seq{\Theta}$:
  \begin{eqnarray}
  \lefteqn{\nabla_{\seq{\Theta}} J(\seq{\Theta})}&& \nonumber\\
  & \!\!\!\!\!\!= & \!\!\!\!\!\nabla_{\seq{\Theta}} \sum_{z \in \seq{Z}} p(z|\seq{x}) \mathcal{L}(\{\seq{y}\}| \seq{x}, z) \nonumber\\
   &\!\!\!\!\!\! = & \!\!\!\!\! \sum_{z \in \seq{Z}} [\mathcal{L}(\{\seq{y}\}| \seq{x}, z) \nabla_{\seq{W}} p(z|\seq{x}) + p(z|\seq{x}) \nabla_{\seq{G}} L(\{\seq{y}\}| \seq{x}, z)]  \nonumber\\
  &\!\!\!\!\!\!= & \!\!\!\!\!\! E_{p(z|\seq{x};\seq{W})} [ \mathcal{L}(\{\seq{y}\}| \seq{x}, z) \nabla_{\seq{W}} \log p(z|\seq{x}) \!+\! \nabla_{\seq{G}} \mathcal{L}(\{\seq{y}\}| \seq{x}, z)]. \label{eq:obj} \nonumber
 \end{eqnarray}

\noindent 
A major difficulty faced by our model is the intractability in 
calculating the above gradient since the expectation involves a huge number of terms for a large latent space $\seq{Z}$.
To address this problem, we propose to use reinforcement learning for optimization, which is shown in Algorithm~\ref{fig:rl}.

In Step~1, 
we use existing keyword extraction tools to extract keywords for each post as the latent words to pre-train the latent word inference network. 
Then we use the top one inferred word from the latent word inference network as the input $z$ to pre-train the generation network with the standard loss function Eq.~\ref{eq:sum} used in Seq2seq models. 
During training, the REINFORCE algorithm~\cite{Williams1992Simple} is used to train the latent word inference network, and standard back-propagation is adopted for updating the generation network. 

Two key issues remains when we perform the joint training.
First, when
we perform $K$ trials to sample multiple $z$'s,
it may be difficult and time-consuming for the algorithm to explore the large latent space, if we directly use the sampling distribution in Eq.~\ref{eq:softmax}.
Second, the loss function in Eq.~\ref{eq:sum}/Eq.~\ref{eq:min} is not bounded and thus will introduce a large variance in policy gradients, which affects the stability of the algorithm.

To address the first issue, we design a two-step sampling scheme, which can effectively limit the word sampling in a much smaller latent space. For the second issue,  we propose to replace $\mathcal{L}(\{\seq{y}\}| \seq{x}, \seq{z})$ in the policy gradient with a bounded reward $\mathcal{R}(\{\seq{y}\} | \seq{z}, \seq{x})$.


 \begin{algorithm}[tb]
\caption{RL Training Algorithm}
\label{alg}
\begin{algorithmic}[1]
\REQUIRE $(\seq{x}, \{\seq{y}\})$, $\seq{Z}$
\ENSURE $\seq{\Theta}$
\STATE $\seq{\Theta} \leftarrow $ Pre-trained from baseline methods.
\FOR{each $(\seq{x}, \{\seq{y}\})$ }
\FOR{$k=1, \ldots, K$}
\STATE Sample a word $z$ according to the proposed sampling scheme.
\STATE
Obtain the generated sentence
$\hat{\seq{y}} = \mbox{argmin}_{\seq{y}^*} \ell(\seq{y}^* | \seq{x}, z)$.
\STATE Compute the reward according to $\mathcal{R}(\{\seq{y}\}| \seq{x}, z)$.
\STATE Update the latent word inference network via the policy gradient $\mathcal{R}(\{\seq{y}\}| \seq{x}, z) \nabla_{\seq{W}} \log p(\seq{z}|\seq{x})$.
\STATE Update the generation network via the gradient $\nabla_{\seq{G}} \mathcal{L}(\{\seq{y}\}| \seq{x}, z)$.
\ENDFOR
\ENDFOR
\end{algorithmic}
\end{algorithm}

\noindent
\textbf{Sampling multiple diverse latent words}
Firstly, we notice that each input post $\seq{x}$ is often relevant to only a few $z$'s, since the information provided by a short text should be very limited. 
Thus, we maintain a small candidate set $\seq{Z}_{\seq{x}}$ for each $\seq{x}$,
 and only sample the latent $z$ in $\seq{Z}_{\seq{x}}$ using their $p(z | \seq{x}; \seq{W})$.
In our experiment, the candidate set $\seq{Z}_{\seq{x}}$ is constructed using outputs from existing keyword extraction methods on the input $\seq{x}$ and its output sequences $\{\seq{y}\}$. 
\footnote{We have tried using all words in $\seq{Z}$ as $\seq{Z}_{\seq{x}}$ and the performance is much worse. }

Second, it is desirable that $K$ diverse $z$'s can be selected out for each $\seq{x}$. Thus we perform clustering
on the candidate set $\seq{Z}_{\seq{x}}$ using their context vector $\seq{c}_{z}$'s in Eq.~\ref{eq:attn_2}.  For $z$'s in different clusters, they tend to have different attentions on the input post. For each trial of sampling, we only select latent words from clusters, from which no words has been selected before.
 
\noindent\textbf{Reward function design}
A few recent works have been proposed to learn a discriminator for a data-driven reward function~\cite{li2017adversarial,li2017paraphrase}. However, these methods involve more model modules and complex parameter tuning.
To keep our model simple,
 we make use of 
 the generated sentence $\hat{\seq{y}}$ from the generation network (Step~5 in Algorithm~1), and compute the reward with respect to a single $\seq{y}$  using the parameter-free F1 score between $\seq{y}$ and $\hat{\seq{y}}$ which measures the average overlap by treating the two sequences as bags of tokens:
\begin{eqnarray*}
\mathcal{R}(\seq{y} | \seq{z}, \seq{x}) = \left\{
\begin{array}{ll}
\mbox{F1}(\seq{y}, \hat{\seq{y}}), &  \seq{y} \cap \hat{\seq{y}} \neq \emptyset, \nonumber\\
-1, & \mbox{else.}
\end{array}\right.
\end{eqnarray*}
 The reward with respect to $\{\seq{y}\}$ is computed as:
\begin{eqnarray*}
\mathcal{R}(\{\seq{y}\} | \seq{x}, z)  =  \left\{
\begin{array}{ll}
\frac{1}{|\{\seq{y}\}|}\sum_{\seq{y}}
\mathcal{R}(\seq{y} | \seq{x}, z), &\mbox{Eq.~\ref{eq:sum} is used,}\\
 \max_{\seq{y}} \mathcal{R}(\seq{y} | \seq{x}, z), & \mbox{Eq.~\ref{eq:min} is used.} \label{eq:reward}
\end{array}\right.
\end{eqnarray*}
Experimental results show that our model with the use of this simple reward function works very well.

\subsection{Generating Multiple Responses}
With a trained model, we discuss how to generate $K$ diverse responses for a testing input $\seq{x}$.
We set the top 1000 $z$'s of $p(z | \seq{x})$ from the latent word inference network as the candidate set.
Next, we perform clustering on the candidate set similarly as in training and select out the centroid $z$'s of the largest $K$ clusters. 
Finally, each of the selected $z$'s is combined
$\seq{x}$ to generate the response.


\section{Experiments}
In this experiment, we first present our two datasets and introduce the compared methods. The training details including the keyword extraction, clustering and network configurations, are provided in Appendix. 
Next, we evaluate the performance of the various compared methods. 
We further show some analyses to validate the effectiveness of our RL training algorithm.

\subsection{Datasets}
We perform experiments on two short-text conversation tasks in which each input post is generally associated with multiple output sequences.
{\bf Weibo:} 
We used the benchmark dataset~\cite{shang2015neural} and pre-processed it for high-quality data pairs. 
In total, we have above 4 million training pairs.
{\bf Twitter:} We crawled post-response pairs using the Twitter API. With several data cleaning steps, we have around 750 thousand training pairs.
%
Both datasets have the vocabulary size 50,000. Details of the data pre-processing and statistics are in Appendix.

\subsection{Compared Methods
}
{\bf Beam Search(BS)}: We use the vanilla Seq2seq model with attention employed in decoding~\cite{bahdanau2015neural}.
Standard beam search is applied in testing.\\
{\bf Diverse Beam Search(DBS)}~\cite{li2016simple}: We use the same Seq2seq model for training.
In testing, a modified ranking score is used in beam search to encourage diverse results. We set a fixed diversity rate as 0.5.\\
{\bf Maximum Mutual Information(MMI)}~\cite{li2016simple}: We generate the N-best list using the same Seq2Seq model, and then rerank the results by this method. \\
{\bf Multiple-Mechanism(MultiMech)}~\cite{zhou2017mechanism} : 
It introduced latent embeddings for diverse response generation.
Following their setting, we use 4 latent mechanisms.
Then we select top 3 mechanisms and perform beam search to obtain one response for each top mechanism.\\
{\bf CVAE}~\cite{zhao2017learning}: As stated in Section~\ref{sec:related}, it adopted CVAE for diverse response generation. We replace the dialogue acts used in their original model as the keyword used in our pre-trained step.\\
{\bf HGFU}~\cite{yao2017towards}: This model used the PMI to extract a cue word for a post and then incorporate it for generating a more informative response. However, we find calculating the PMI is very time-consuming for a testing sample, which involves $O(\mbox{the sequence length} \times \mbox{vocabulary size} \times \mbox{the number of the training data})$ operations. Thus we change to apply the keyword used in our pre-trained step.\\ 
{\bf Ours}:  We explore our model using the pre-trained networks only~({\bf Ours(Pretrain)})
and two variants utilizing the two loss functions in Eq.~\ref{eq:sum} and~\ref{eq:min} in the generation network, denoted as {\bf Ours(Avg)} and {\bf Ours(Min)}. 

\subsection{Evaluation}
Top 3 results of each method are collected for evaluation.
For automatic evaluations, we report:\\
{\bf BLEU}~\cite{papineni2002bleu}:
A widely used metric to measure the response relevance;\\ 
{\bf Distinct-1/2}: Number of distinct unigrams/bigrams in the generated multiple responses scaled by the total number of generated unigrams/bigrams for each post, which can be considered an automatic metric for evaluating the diversity among the multiple responses generated for a single post.\\
Since automatic metrics for open-domain generative models may not be consistent with human perceptions~\cite{liu2016not}, we also do human annotations to further evaluate the generated results.
We randomly select 200 testing posts for each dataset and recruit 3 annotators from a professional labeling company 
to label the following two aspects for results of all compared methods:\\
{\bf Quality}: Each post-response pair is evaluated with: bad (ungrammatical or irrelevant), normal (basically grammatical and relevant) or good (grammatical and highly relevant with specific meaning and information); responses on normal and good levels are treated as ``acceptable'';\\
{\bf Diversity}: Number of fundamentally distinct responses among the output responses of each method;
thus the largest diversity score of a post is 3 in our setting. 

\subsection{Results on Weibo}

  \begin{table*}[htbp]
 \small
	\begin{center}
     \setlength{\tabcolsep}{3mm}
			\begin{tabular}{ l | c| c | c | c | c | c}
            \hline
				Method & BLEU-1 & BLEU-2 & BLEU-3 & BLEU-4 & distinct-1 &distinct-2\\
				\hline
			     BS & 24.784& 18.923& 13.362& $7.866$ &$44.316$&$53.250$\\
				DBS & 24.187& 18.437& 13.016& $7.674$ &$44.962$&$54.114$\\
                MMI & 25.683& 19.597& 13.845& $8.187$ &$43.369$&$52.265$\\
                CVAE &26.575 & 18.290& 12.302& $6.510$ & $38.158$ &$47.272$ \\
				MultiMech & 24.436& 18.952& 13.644& $8.403$ & $41.321$ &$49.592$ \\
         		HGFU & \bf{43.530} & \bf{32.356} & \bf{22.826} & \bf{14.043} & $36.262$ &$42.323$ \\

                \hline
				Ours(Pretrain) & 31.135$^*$ & 23.361 & 16.889 & 10.594 & 58.534$^*$ & 76.023$^*$\\
                Ours(Avg) & 30.435& 23.187& 16.511&$10.086$ &$43.618$ & $51.769$  \\
                Ours(Min) & 31.067 & 23.523$^*$ & 16.893$^*$ & 10.945$^*$ & \bf{62.407} & \bf{76.767}\\\hline
			\end{tabular}
		\caption{Automatic evaluation results on Weibo. The best/second best results are bold/starred.}
\label{tab:weibo_auto}
	\end{center}
\end{table*}

\begin{figure*}[htbp]
\begin{center}
 \includegraphics[width=0.7\linewidth]{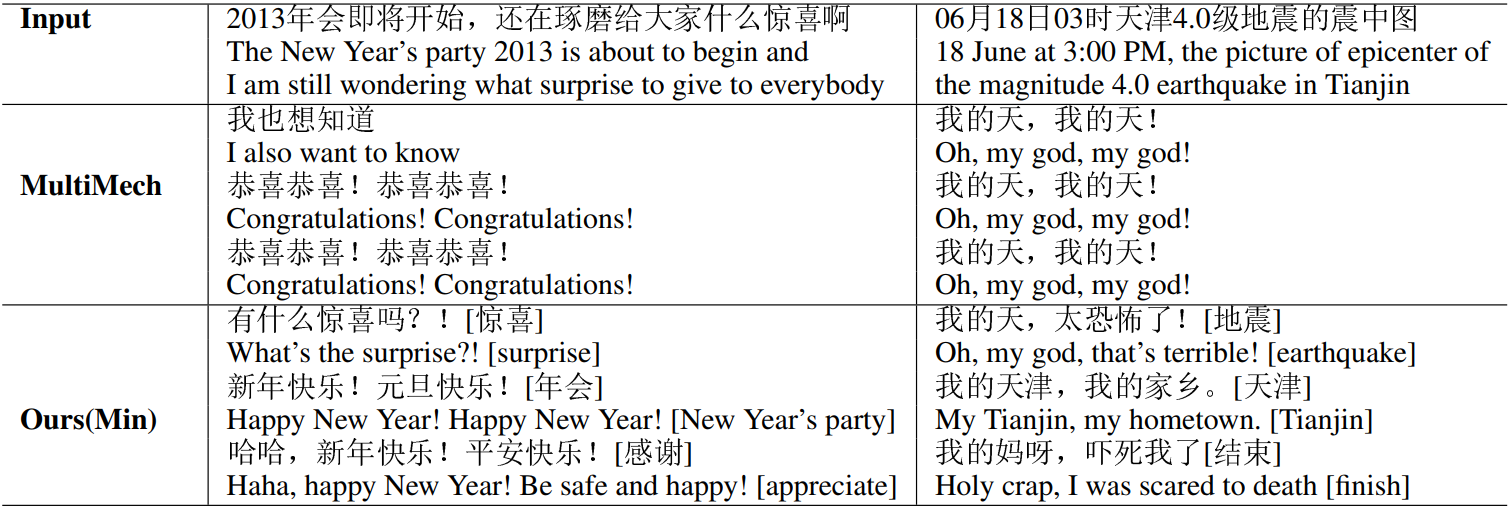}
\end{center}
\caption{Examples showing the generated responses on Weibo. Words in the brackets are the selected latent words.}
\label{fig:weibo_case}
\end{figure*}

 \begin{table}[htbp]
 \small
	  \setlength{\tabcolsep}{2.5mm}
	\begin{center}
			\begin{tabular}{ l | c| c | c}
                \hline
				\multirow{2}{*}{Method} & \multicolumn{2}{c|}{Quality}   & \multirow{2}{*}{Diversity} \\\cline{2-3}
				&  \%Acceptable & \%Good & \\
				\hline
			     BS & $44.611$ & $4.333$ & $0.678$\\
				DBS & $44.667$ & $4.389$ & $0.683$\\
                MMI & $44.611$ & $4.056$ & $0.680$\\
                CVAE & $ 29.833 $ & $ 1.888 $ & $ 0.380$\\
				
				MultiMech & $ 60.388 $ & $ 5.000 $ & $ 0.806$\\
				HGFU & $ 60.833 $ & $ 7.111 $ & $ 0.846$\\
                \hline
				Ours(Pretrain) & $46.722$ & $9.611$ & $1.148$\\
                Ours(Avg) & $49.278$ & $8.556$ & $0.843$\\
                Ours(Min) & $\textbf{63.000}$ & $\bf{20.944}$ & $\bf{1.436}$\\\hline
			\end{tabular}
		\caption{Human evaluation results on Weibo.}
\label{tab:weibo_human}
	\end{center}
\end{table}

\textbf{Overall Performance}
Results of automatic evaluations and human annotations are shown in Table~\ref{tab:weibo_auto} and~\ref{tab:weibo_human} respectively. 
%
In terms of BLEU, HGFU performs the best, while Ours(Min) achieves the second best on BLEU-2/3/4. However, BLEU only evaluates how much overlapped words the generated responses have towards the given references.
In open-domain conversations, responses not close to the references may not necessarily be poor. Thus, the quality scores from the human annotations are more reliable. As shown, Ours(Min) obtains the highest acceptable ratio. More importantly, our model achieves a much higher good ratio, with approximately 195\% improvement over HGFU. This indicates that our method can output more high-quality responses.
%
In terms of distinct-1/2 and diversity, Ours(Min) performs the best. Especially from the human evaluation results, Ours(Min) obtains about 110\% improvement over the Seq2seq baselines (BS, DBS and MMI), as well as 70\% over the enhanced methods (MultiMech and HGFU). 
This validates that our method can output more diverse responses while maintaining each of them to be relevant to the post.

Comparing with Ours(Pretrain), Ours(Avg) cannot obtain consistent improvement on both quality and diversity. This shows that the loss in Eq.~\ref{eq:min} is more suitable for our problem setting as discussed in Section~3.

We notice that the re-ranking methods (DBS and MMI) do not outperform BS considering both quality and diversity. As stated in introduction, these methods may have limited effectiveness by only encouraging responses with different words to be ranked into higher positions during testing. Also, the diversity of CVAE is much worse than the other methods. This is consistent with our discussion in Section~\ref{sec:related} that directly sampling from a fixed Gaussian prior tends to cause mode collapse and yields little response diversity.


 \begin{figure}[htbp]
\begin{center}
\subfigure[]{\label{fig:latent_word}
 \includegraphics[width=0.43\linewidth]{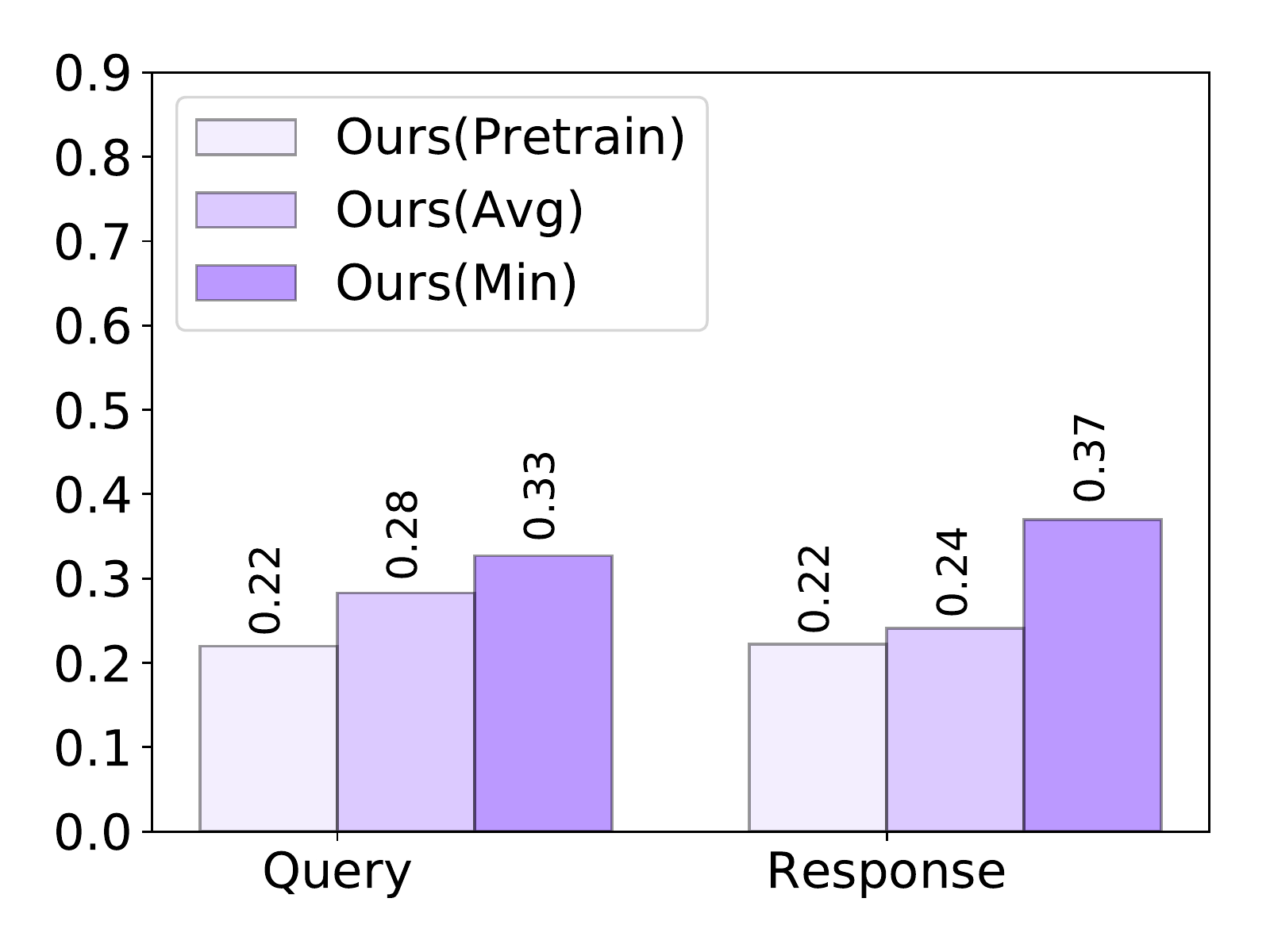}}
 \subfigure[]{\label{fig:attention}
  \includegraphics[width=0.51\linewidth]{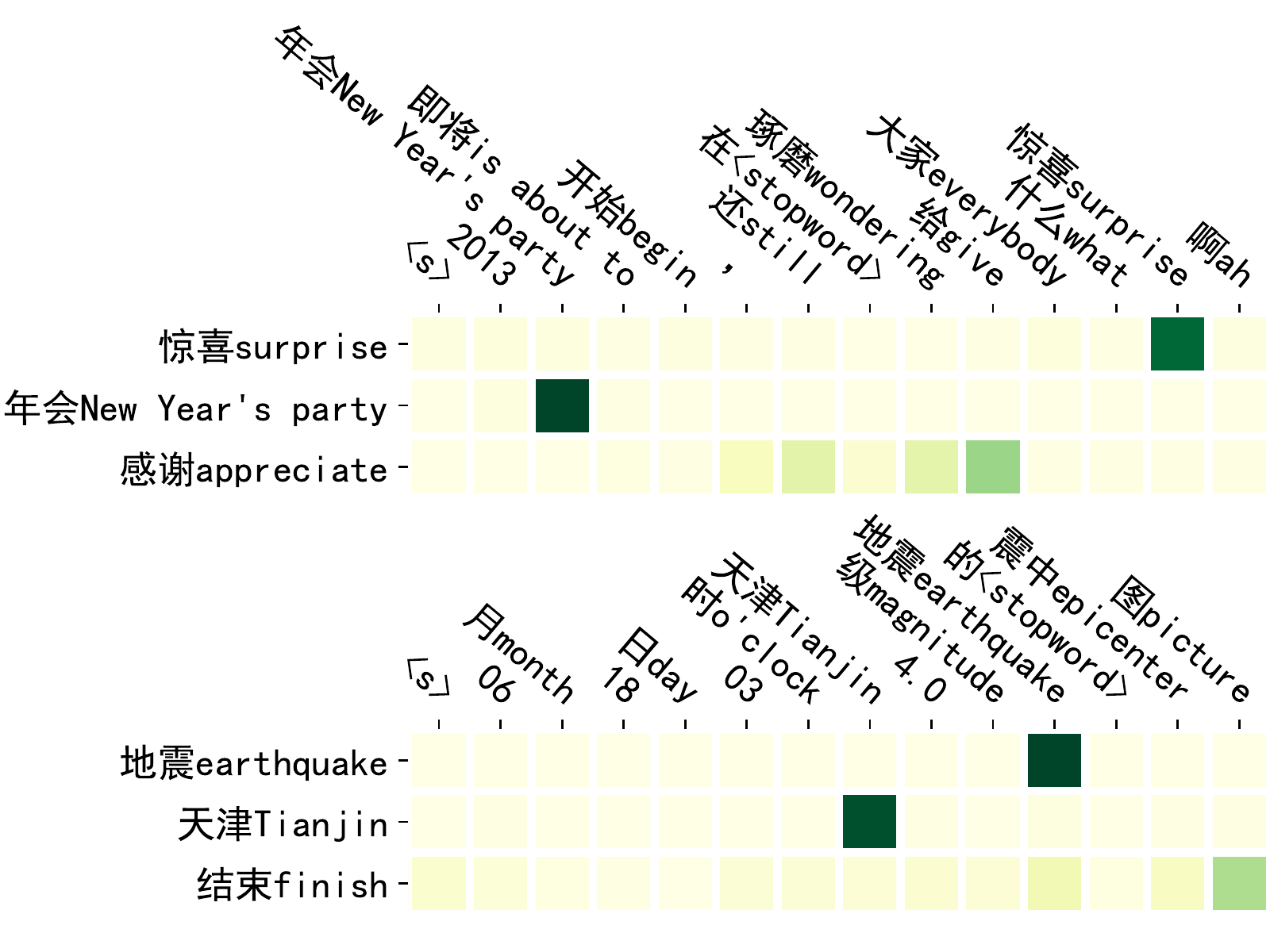}
 }
\end{center}
\caption{
(a) Cosine similarity between the latent word and the query/response. 
(b) Visualization of the attention paid to the relevant parts of the input post in Fig.~\ref{fig:weibo_case} for each of the selected words; a large version of this figure is shown in Appendix for better readability.}
\end{figure}


  \begin{table*}[htbp]
 \small
	\begin{center}
    \setlength{\tabcolsep}{3mm}
			\begin{tabular}{ l | c | c | c | c | c | c}
            \hline
				Method & BLEU-1 & BLEU-2 & BLEU-3 & BLEU-4 &distinct-1 &distinct-2\\
				\hline
			     BS & 39.595& 34.331& 27.727& $22.857$ &$51.867$&$63.951$\\
				DBS &37.004 & 32.075& 25.836& $21.198$ &$53.703$&$66.226$\\
                MMI & 44.306& 38.727& 31.760& $26.661$ &$48.678$&$60.135$\\
				 CVAE & 35.071& 29.487& 22.509& $17.380$ &$52.451$ &$64.876$\\
                MultiMech & 75.164& 50.679& 30.897& $19.434$ &$42.717$ &$48.153$\\    
                HGFU & 28.145& 18.214& 12.963& $9.488$ &$5.786$ &$6.609$\\

                \hline
				Ours(Pretrain) & 82.751& 64.902& 47.177& $34.282$ &$52.281$&$66.642$\\
                Ours(Avg) & $82.886^*$ & $66.586^*$ & $48.288^*$ &$34.891^*$ &$\bf{55.371}$ & $67.210^*$  \\
                Ours(Min) & \textbf{86.682} & \textbf{74.356} & \textbf{60.294} & $\bf{49.965}$ &$53.904^*$ & $\bf{67.301}$\\\hline
			\end{tabular}
		\caption{Automatic evaluation results on Twitter. The best/second best results are bold/starred.}
\label{tab:twitter_auto}
	\end{center}
\end{table*}

 \begin{table}[htbp]
 \small
	  \setlength{\tabcolsep}{2.5mm}
	\begin{center}
			\begin{tabular}{ l | c| c | c}
            \hline
				\multirow{2}{*}{Method} & \multicolumn{2}{c|}{Quality}   & \multirow{2}{*}{Diversity} \\\cline{2-3}
				&  \%Acceptable & \%Good & \\
				\hline
			     BS & $28.055$ & $3.778$ & $0.657$\\
				DBS & $29.000$ & $3.722$ & $0.697$\\
                MMI & $31.667$ & $4.667$ & $0.678$\\
				CVAE & $ 20.666 $ & $ 3.055 $ & $ 0.475$\\
				
				MultiMech & $ 21.166 $ & $ 3.500 $ & $ 0.536$\\
				HGFU & $ 14.000 $ & $ 2.722 $ & $ 0.245$\\
                \hline
				Ours(Pretrain) & $27.389$ & $4.333$ & $0.585$\\
                Ours(Avg) & $30.667$ & $4.444$ & $0.570$\\
                Ours(Min) & $\bf{42.000}$ & $\bf{7.944}$ & $\bf{0.713}$\\\hline
			\end{tabular}
		\caption{Human evaluation results on Twitter.}
\label{tab:twitter_human}
	\end{center}
\end{table}	

\begin{figure*}[htbp]
\begin{center}
 \includegraphics[width=0.68\linewidth]{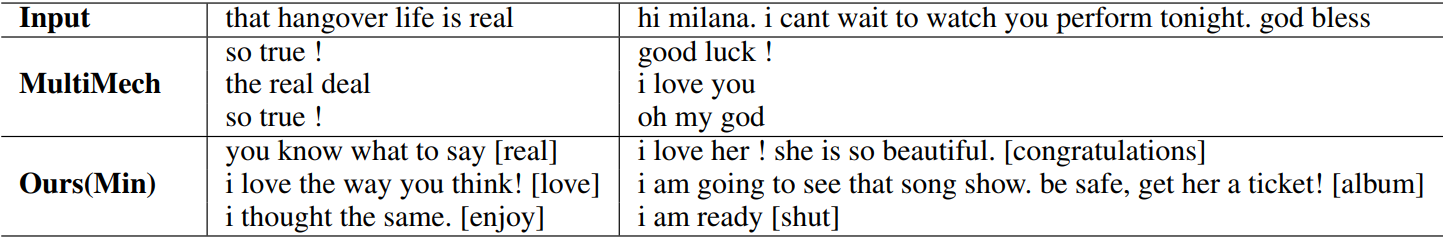}
\end{center}
\caption{Examples showing the generated responses on Twitter. Words in the brackets are the selected latent words.}
\label{fig:twitter_case}
\end{figure*}

\noindent \textbf{Analysis and Case study}
We first evaluate the overall relevance between the latent words selected by three variants of our methods and their corresponding post/generated response. We use a simple method to obtain the sentence embeddings of the posts/ generated responses, which takes the average of all the word embeddings in the sentences~\cite{adi2016fine,zhao2018unsupervised}. We use pre-trained word embeddings trained with the C-BOW model~\cite{mikolov2013efficient} on the Chinese Wikipedia corpus. Then we compute the cosine similarity between the latent word and the sentence embedding, and normalize the score into [0,1]. Results are shown in Fig.~\ref{fig:latent_word}. As can be seen, both Ours(Avg) and Ours(Min) can select out latent words that are more relevant to both the posts and responses than Ours(Pretrain). Meanwhile, the improvement of Ours(Min) is much larger than that of Ours(Avg).

Fig.~\ref{fig:weibo_case} illustrates two example posts with their generated responses. \footnote{Full results of all methods are shown in Appendix.} As can be seen, the responses with different mechanisms can be identical. The reason may be that the number of mechanisms is very small and each mechanism can only capture high-level responding characteristics shared by many input data. A given post may not be sensitive to different responding mechanisms so that the same response will be generated.  Whereas, the responses by our methods are driven by different latent words, which contain more specific data-dependent information.  If the selected words have far different meanings, the generated responses are highly probable to be different.

We further examine the relationship between the latent words selected in our method and the generated results.
Fig.~\ref{fig:attention} shows the attention vector of each selected word over its corresponding post.
The first two latent words for both posts are highly aligned to one word in the post, which is often the case that this word appears in the post. Then our method probably generates specific responses about that word.
If the word is aligned with multiple words as the third one for both posts, the generated response
is relevant to the these attended words but not the latent word itself.
Overall, since we select words from different clusters, which are computed using the context vector $\seq{c}_{z}$, our method tends to select words attended on different parts of the post and the generated responses are generally with diverse meanings. 

\subsection{Results on Twitter}

 Table~\ref{tab:twitter_auto} shows the automatic evaluation annotation results on Twitter. On all metrics, our proposed
 models outperform the other compared methods.
 The human evaluation results are provided in Table~\ref{tab:twitter_human}, and our method is still the best among all compared methods.
 Especially on the the Good ratio, Ours(Min) has an increase of 70\% over MMI (the best re-ranking method) and 126\% over MultiMech (the best diversity-promoting method). 
 Generated responses of two posts are provided in Fig.~\ref{fig:twitter_case}. 
 Full results of all methods are again provided in Appendix.

\subsection{Analysis of RL Training}

\begin{figure}[htbp]
\begin{center}
\subfigure[]{\label{fig:latent}
 \includegraphics[width=0.45\linewidth]{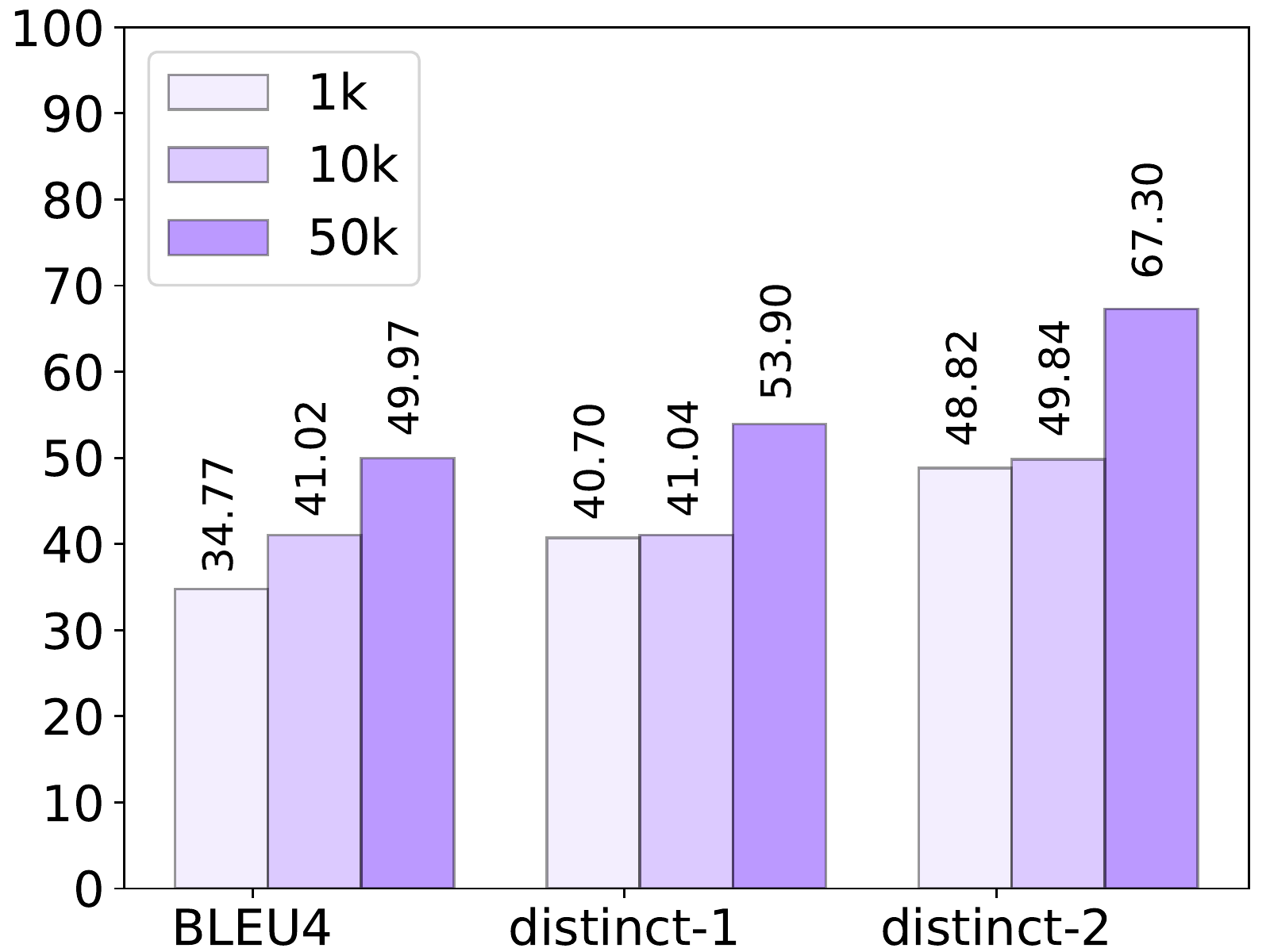}}
 \subfigure[]{\label{fig:reward}
  \includegraphics[width=0.45\linewidth]{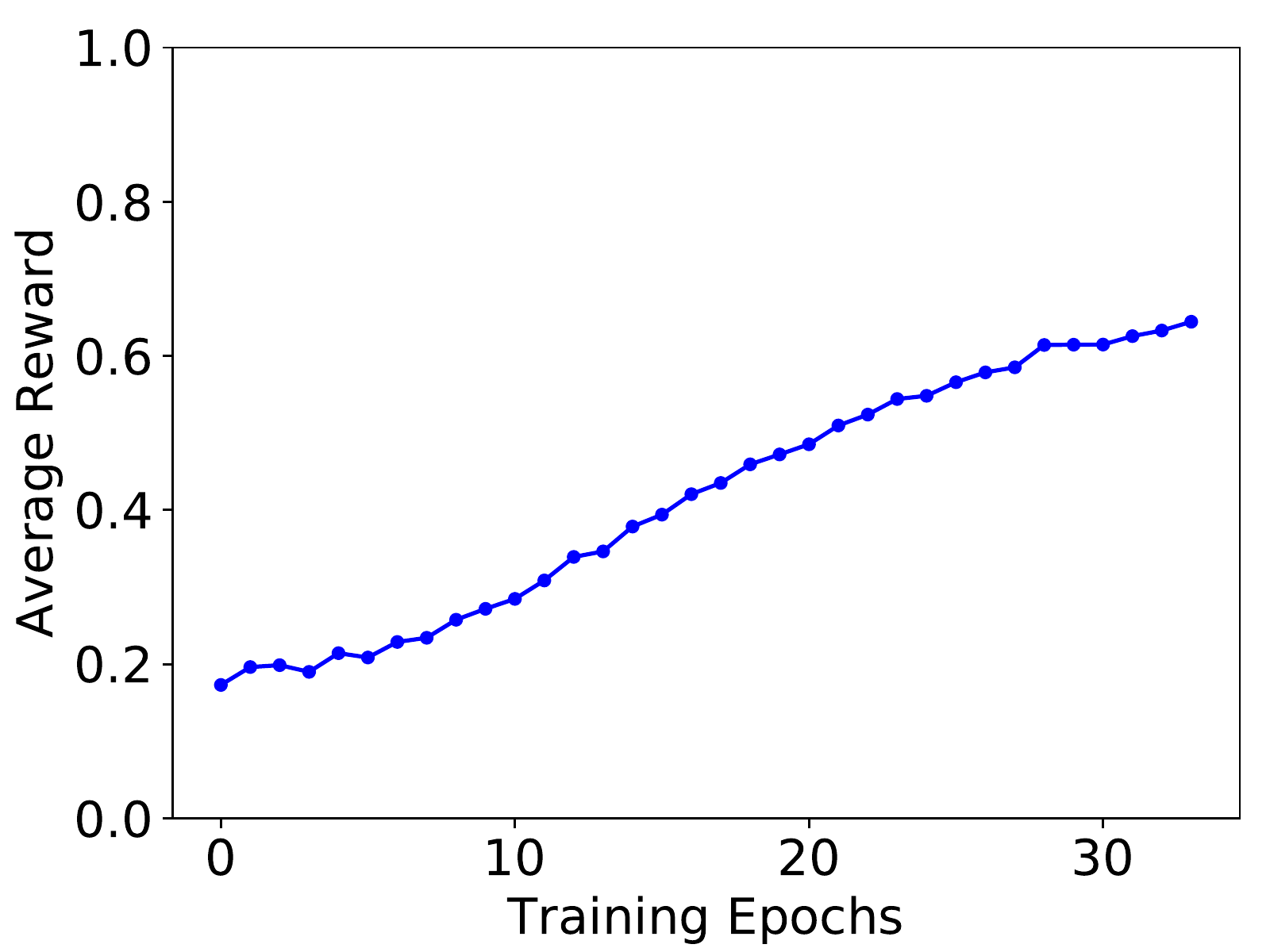}
 }
\end{center}
\caption{(a) BLEU and distinct-1/2 with the size of latent space as 1k, 10k and 50k. (b) Training rewards with different epochs.}
\end{figure}

We vary the size of the latent space on Twitter to test if a large latent space is needed in our model. 
In Fig.~\ref{fig:latent}, we show the results of Ours(min) with the latent space using the top 1k, 10k and all words in the vocabulary.
We find out that all metric scores decrease drastically with the use of a smaller latent space.
Thus, it is critical to use a large latent space for our model to perform well.

In Fig.~\ref{fig:reward}, we plot
the average training reward of Ours(Min) obtained by each epoch on Twitter.
The reward increases as the training algorithm iterates. It indicates our designed RL training algorithm is capable of revising its policy to obtain more accurate latent variable distributions to generate better responses.
 
\section{Conclusion}

 We propose a new response generation model that learns the post with its set of responses jointly for short-text conversation.
 Our model can be optimized in a reinforcement learning algorithm, which can deal with the large latent space assumed in our model. 
 By sampling multiple diverse latent words from the latent word inference network, the generation network can output different responses.
We perform extensive experiments on two real datasets collected from Weibo and Twitter.
Results show that, our method
can effectively increase both the quality and diversity of the multiple generated responses compared with existing baselines and several state-of-the-art generative methods. 
\section*{Acknowledgements}
This work was supported by National Natural Science Foundation of China (Grant No. 61876120).
\bibliography{main}

\appendix
\section{Data Preprocess}
For {\bf Weibo}, we perform the following pre-processing steps:
\begin{itemize}
	\item Hash tags and some special tokens such as ``Zhuan~(forward)" in the post/response are removed;
	\item Irrelevant pairs are removed, which are detected by a Twitter LDA model~\cite{Zhao2011Comparing}. We use a subset of posts to train the LDA model and each post/response is represented with its topic-distribution vector.
	By computing the cosine similarity between the post and the response
	with their topic-distribution vectors, we filter out those pairs with similarities lower than a threshold.
	
	\item To handle the out-of-vocabulary (OOV) problem, we keep the top 50,000 most frequent words and set the rest as UNK.
	
\end{itemize}
For {\bf Twitter}, we perform the following data cleaning steps:

\begin{itemize}
	\item Non-ASCII characters except spaces and periods are removed; 
	\item All the hash tags and links of a tweet are removed;
	\item Tweets with less than 10 characters are removed;
\end{itemize}

\begin{table}[htbp]
	\small
	\begin{center}
		\begin{tabular}{ l| c| c}

			& Weibo & Twitter \\
			\hline
			\#training $(\seq{x},\seq{y})$ & 4,245,992 & 754,428 \\
			\#training $(\seq{x},\{\seq{y}\})$ & 211,208   & 6000\\
			\#valid $(\seq{x},\seq{y})$ & 3,200  & 3,200 \\
			\#valid $(\seq{x},\{\seq{y}\})$ &  800 & 800\\	
			\#testing $(\seq{x},\seq{y})$ & 3,200  & 3,200 \\
			\#testing $(\seq{x},\{\seq{y}\})$ &  800 & 800\\	

		\end{tabular}
		\caption{Statistics of datasets used.}
		\label{tab:stat}
	\end{center}
\end{table}	

\noindent
We extract keywords used in the pre-trained steps in our algorithm from a post-response pair using Jieba for our Chinese dataset {\bf Weibo} \footnote{~\url{https://github.com/fxsjy/jieba}} and RAKE \footnote{~\url{https://github.com/aneesha/RAKE}} for the English dataset {\bf Twitter} respectively.

\section{Implementation Details}
We implemented our methods using PyTorch with the following parameters (according to the loss on the validation set):
for the latent word inference network, the dimension of word embedding, hidden size of the GRU encoder for each direction and FC layer is 620, 500, and 1000 respectively.

For the generation network, we use the same parameter setting of the word embedding and GRU encoder as in the latent word inference network, but their parameters are not shared. The latent word $z$ has its representation of size 1000. The response decoder has a hidden size of 1000.

All parameters are initialized from a uniform distribution between -0.1 and 0.1.
The models are trained using Adam~\cite{kingma2014adam} with the learning rate of 0.0002 and batch size of 256. 

We sample K diverse $z$'s using K-Means technique. Specifically, we initialize the K-Means algorithm with K=10 and then sample one $z$ per cluster.
After this we test our models and the other models with beam size 3 and 100 respectively.

\section{Additional Results}
\begin{figure}[htbp]
	\begin{center}
		\subfigure[]{\label{fig:reward}
			\includegraphics[width=\linewidth]{weibo_attn.pdf}
		}
		\subfigure[]{\label{fig:reward}
			\includegraphics[width=\linewidth]{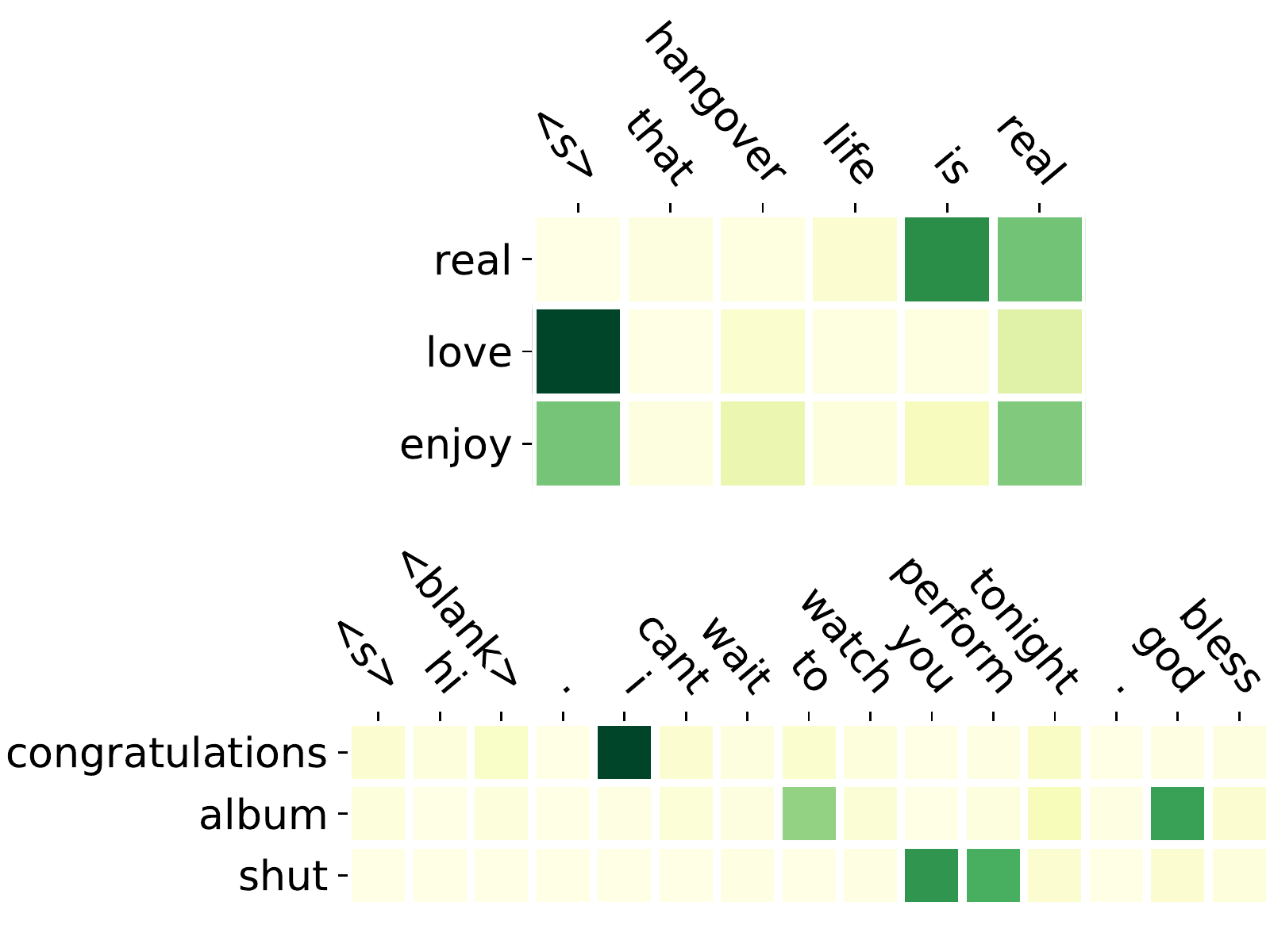}
		}
		\\
	\end{center}
	
	\caption{(a) A large version of Fig 3 in the paper. (b) Visualization of the attention paid to the relevant parts of the input post in Fig.~\ref{fig:twitter_case} in this appendix for each of the selected latent words.}
	\label{fig:a}
\end{figure}

Fig.~\ref{fig:weibo_case}/~\ref{fig:twitter_case} shows the example cases of all our compared methods on Weibo/Twitter respectively.
Fig.~\ref{fig:a} visualizes the attention of the two examples on Twitter.
We can observe similar results as in the examples on Weibo:
If a latent word is highly aligned to only one word in the post, our method tends to generate a specific response about that word.
If the word is aligned with multiple words, the generated response
is relevant to the these attended words but not the latent word itself.

\begin{figure*}[htbp]
	\begin{center}
		\includegraphics[width=0.85\linewidth]{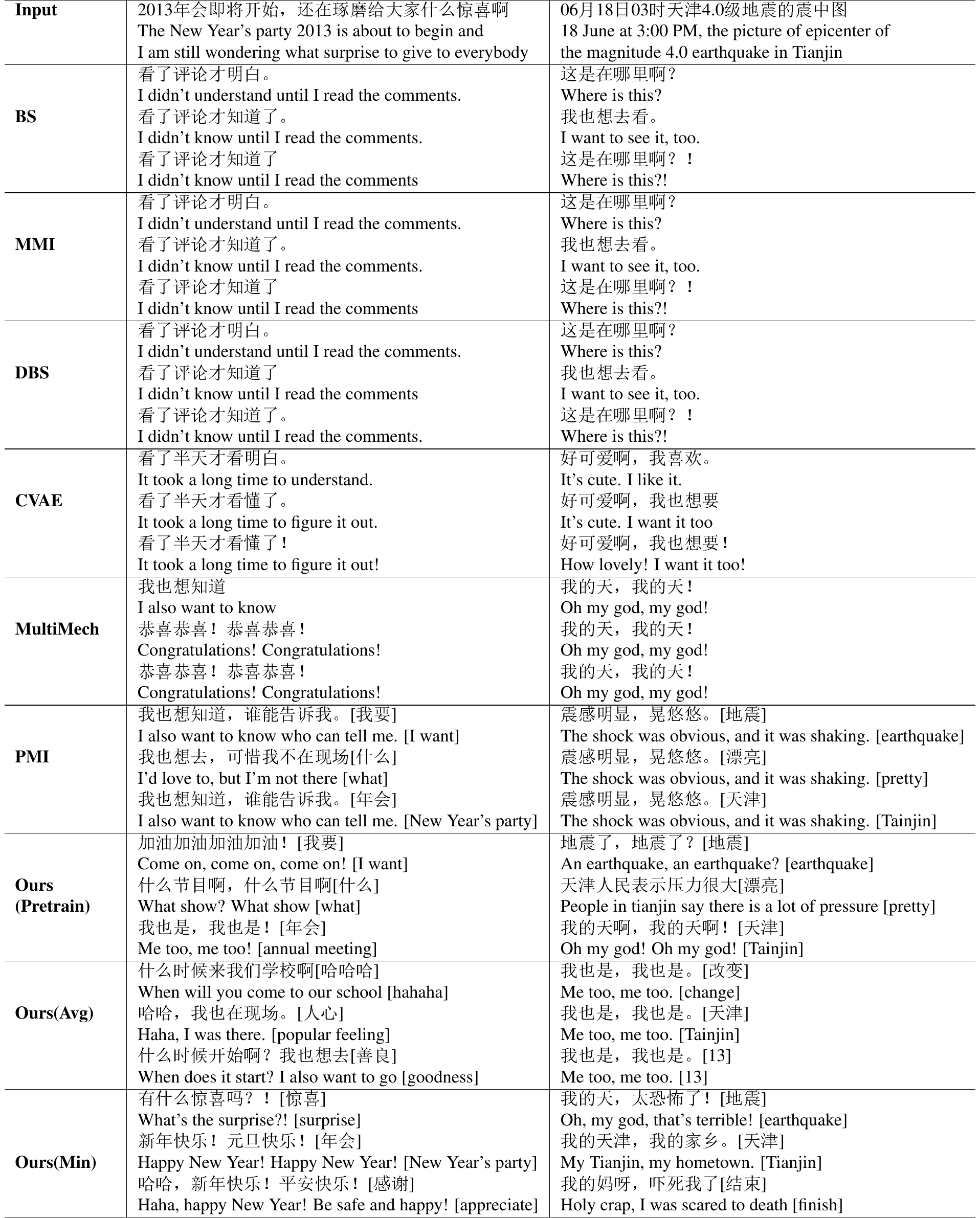}
	\end{center}
	\caption{Examples showing the generated responses on Weibo. Words in the brackets are the selected latent words.}
	\label{fig:weibo_case}
\end{figure*}

\begin{figure*}[htbp]
	\begin{center}
		\includegraphics[width=0.85\linewidth]{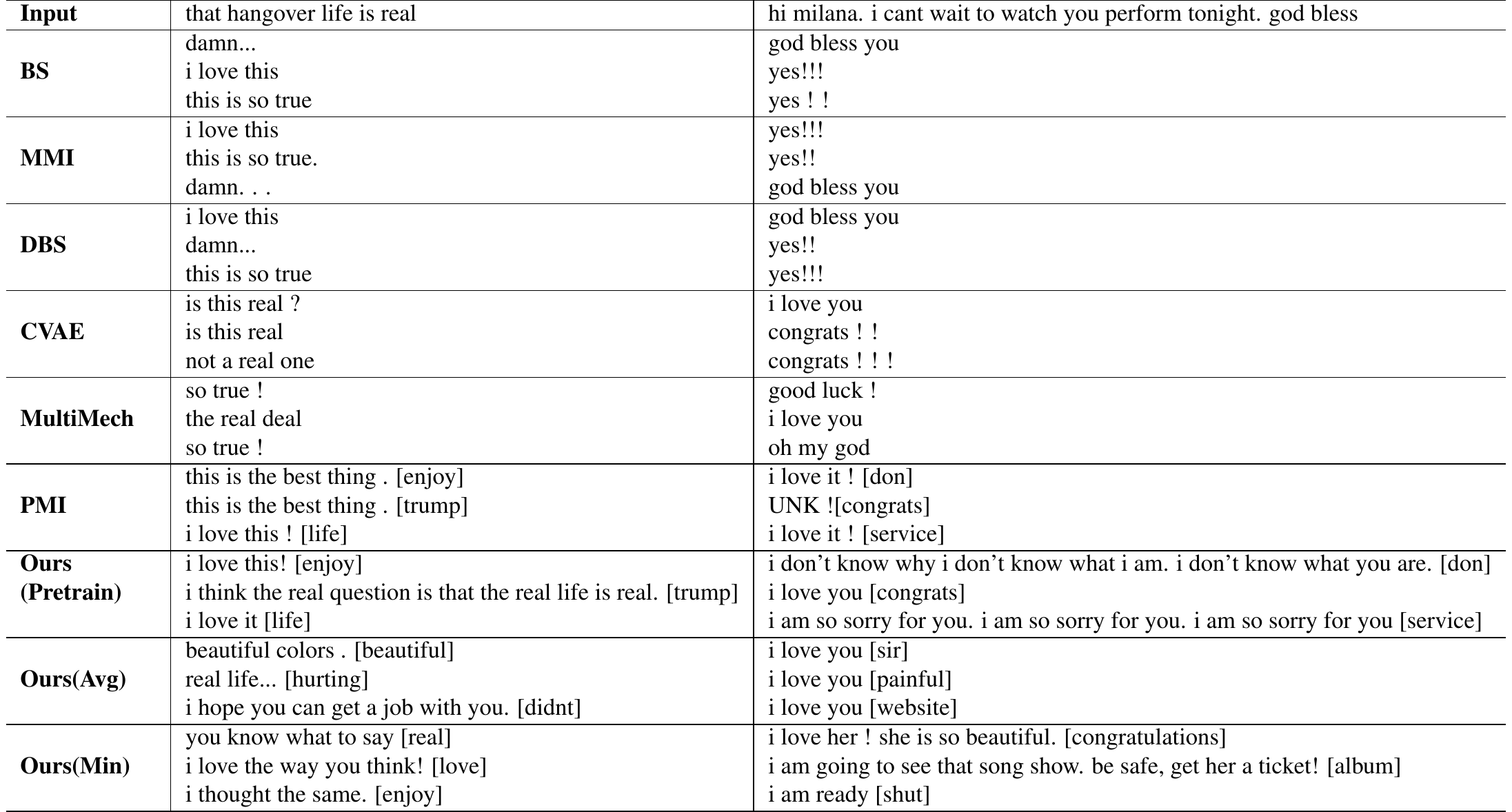}
	\end{center}
	\caption{Examples showing the generated responses on Twitter. Words in the brackets are the selected latent words.}
	\label{fig:twitter_case}
\end{figure*}

\bibliographystyle{aaai}
\end{document}